  \documentclass[preprint]{elsarticle}

   \usepackage{amssymb}

\usepackage{diagbox}
\usepackage{makecell}
\usepackage{array}
\usepackage{color}
\usepackage{subfigure}

\journal{Journal of Visual Communication and Image Representation~}

\begin{document}

\begin{frontmatter}


\title{Instance Search via  Instance Level \\ Segmentation and Feature Representation}


\author[1]{Yu Zhan}
\ead{yeezytaughtme@stu.xmu.edu.cn}
\address[1]{The School of Information Science and Technology, \\
Xiamen University\\
Xiamen, 361005, P. R. China.\\}

\author[1]{Wan-Lei Zhao\corref{cor1}}
\cortext[cor1]{Corresponding author}
\ead{wlzhao@xmu.edu.cn}

\begin{abstract}
Instance search is an interesting task as well as a challenging issue due to the lack of effective feature representation. In this paper, an instance level feature representation built upon fully convolutional instance-aware segmentation is proposed. The feature is ROI-pooled from the segmented instance region. So that instances in various sizes and layouts are represented by deep features in uniform length. This representation is further enhanced by the use of deformable ResNeXt blocks. Superior performance is observed in terms of its distinctiveness and scalability on a challenging evaluation dataset built by ourselves. In addition, the proposed enhancement on the network structure also shows superior performance on the instance segmentation task.
\end{abstract}


\end{frontmatter}


\section{Introduction}
With the proliferation of massive multimedia contents in our daily life, it is desired that users are allowed to browse over relevant images/videos in which the specified visual instance (e.g., an object or a landmark or a person) appears. This is known as instance search~\cite{Awad17}, which arises from several application scenarios such as online product search in the shopping website, video editing, and person re-identification, etc.

Instance search is essentially different from conventional content-based image retrieval (CBIR)~\cite{RuiHC99, AlZubiAR15} in several perspectives. First of all, in instance search, the query is a visual object that is outlined (usually by a bounding box) in an image. While in CBIR,  the whole image is treated as the query. Secondly, instance search requires the intended visual objects to come from the same instance (while possibly under different transformations) as the query~\cite{Awad17}. In contrast, CBIR only requires the returned contents to be visually similar as the query image no matter whether they share the same origin. Moreover, instance search should localize the target instance in the returned images.

There are basically two stages in visual content search system, namely feature representation~\cite{Sivic03, Jegou10, JegouDS10, RetaCMGAD18, DalalT05, PerronninSM10, ZhangLZCY15, RaveauxBO13, WANG201363, Babenko14, Razavian14, Babenko15, NgYD15} and fast retrieval~\cite{JegouDS11, DatarIIM04, MujaL09, MujaL14}. In the whole process, feature representation plays the key role to the success of the system. On one hand, features are required to be robust to various image transformations, such as scaling, rotation and occlusions, motion blur, etc. On the other hand, they should be distinctive enough so that the retrieval quality does not suffer severe degradation as the scale of the reference set grows.

In the existing solutions, instance search has been mainly addressed by conventional approaches that are originally designed for image search~\cite{Awad17, RuiHC99}, such as bag-of-visual words (BoVW)~\cite{Sivic03}, RoI-BoVW~\cite{ZhangLZCY15}, VLAD~\cite{Jegou10} and FV~\cite{PerronninSM10}. All these approaches are built upon image local features such as SIFT~\cite{Lowe04}, RootSIFT~\cite{ArandjelovicZ12}, SURF~\cite{BayTG06}. Although local features are much more distinctive than global features, they are still unsuitable for instance search task. First of all, local features are not robust to out-of-plane rotation and deformation, both of which are widely observed in the real world. Moreover, it is not rare that very few local features are extracted from transparent objects (e.g., bottles) or objects with flat surface (e.g., balls). Additionally, it is not guaranteed that the regions covered by local features are exactly from one instance. As a result, the local features used to describe a target instance are more or less contaminated by the contents from the background. For this reason, similar as global features, isolated feature representation for individual instances is not desirable.  

Recently, pre-trained CNNs are gradually introduced to image retrieval tasks~\cite{Babenko14, Razavian14, Babenko15, NgYD15, ZhengZWWT16, ArandjelovicGTP16, XieZWYT16} due to their great success in visual object classification tasks~\cite{DengDSLL009}. In the existing practices, image features are typically extracted from the whole image or a series of local regions with convolution or fully connected layers. Encouraging results are observed on the landmark retrieval tasks in~\cite{Razavian14, Babenko15}. However, they are unfeasible for instance representation since it is essentially a type of global feature. The feature vector is comprised by a mixture of activations from a variety of latent instances in the image. Although recent research~\cite{Tolias15, Kalantidis16} attempts to localize the representation to regional level, exhaustive sliding search or feature aggregation is still inevitable. Moreover, since such region level representation is given by a coarsely restricted region, their improvement is still limited.

In this paper, an instance level feature representation is proposed, which is based on an effective instance segmentation approach, namely fully convolutional instance-aware semantic segmentation (FCIS)~\cite{Li17}. Individual instances present in the image are detected and segmented on pixel level by FCIS. This is essentially different from the approach presented in~\cite{Salvador16}, in which the segmentation only reaches to the semantic category level. With the instance level segmentation, feature representation of each instance is derived from the feature maps of convolution layers using ROI pooling. So that instances in different sizes and layouts are represented with the feature vectors of the same size. In order to enhance the performance, two modifications have been made on the FCIS network.\

\begin{itemize}
	\item The back-bone network of FCIS is replaced with a more powerful ResNeXt-101~\cite{Xie17} without increasing extra FLOPs complexity or the number of parameters;
	\item To enable the receptive field to be adaptive to the various shape of potential objects, the plain layer in ResNeXt-101's final stage is replaced with deformable convolution~\cite{Dai17}.
\end{itemize}
To the best of our knowledge, this is the first piece of work that visual instances are represented by features derived exactly from the instance region. Moreover, due to the lack of publicly available testing benchmark for instance search, a new dataset called \textit{Instance-160} is constructed by harvesting test videos that are originally used for visual object tracking evaluation.

\section{Framework for Instance Search}
\label{sec:mthd}

\subsection{Instance Level Feature Representation}
\begin{figure}[htbp]
\begin{center}
	\centering
	\includegraphics[width=0.84\linewidth]{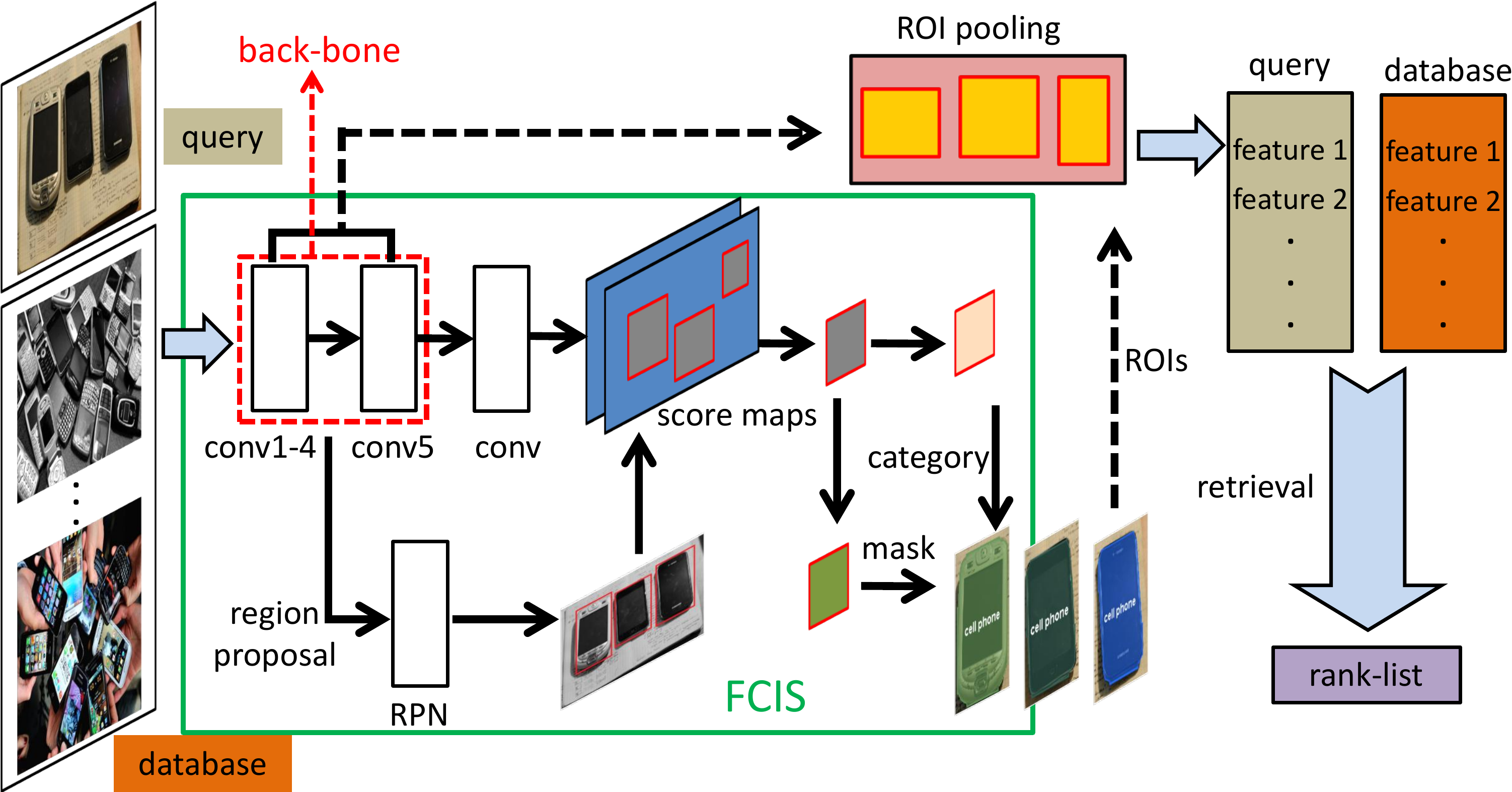}
\end{center}
	\caption{Framework of instance-level feature representation from convolutional activations of FCIS~\cite{Li17}. Processing flow with black arrows and dashed lines denote the proposed modification and enhancement over FCIS.}
	\label{fig:framework}
\end{figure}

Fully convolutional instance-aware semantic segmentation (FCIS)~\cite{Li17} is designed primarily for instance segmentation and detection. The framework of FCIS is given as a sub-figure in Fig.~\ref{fig:framework}, which is inside the bounding box in green. In the network, the idea of ``position-sensitive score map'' is adopted to perform segmentation and detection simultaneously. These two sub-tasks share the same set of score maps by assembling operation according to the region of interest (ROI). ROIs are generated by region proposal network (RPN), which is added on top of ``conv4''. The score maps output ``inside'' and ``outside'' scores for the mask prediction and classification jointly. For details, readers are referred to~\cite{Li17}.

As seen from Fig.~\ref{fig:framework}, there are three outputs from FCIS for one image, namely, the segmented instances (given as instance masks) and the corresponding category label, along with the bounding box of each instance. In order to extract the feature for each segmented instance, another pipeline is introduced into FCIS framework. Namely, with the generated bounding box, ROI pooling is performed on the feature maps that are generated in the convolution stages. This feature extraction pipeline is shown on the up-right of Fig.~\ref{fig:framework}. Since the size of feature map is different from the input image and varies from layer to layer, bounding box of each instance is scaled accordingly to fit the size of the feature map when we perform ROI pooling. The maximum activation is extracted from the scaled ROI region as one dimension of the feature representation. This ROI pooling is applied on all feature maps in the same layer. As a consequence, the size of the output feature equals to the number of the feature maps. Instances in different sizes and layouts are represented with the same size of feature vectors. Since the segmentation is precise and clean, this feature representation is on instance level of real sense. All per-ROI computation is simple and fast with a negligible cost, compared with forward pass. 

Intuitively, convolution layers keep more abstract visual information as network goes deeper. It is therefore widely believed that shallower convolution layers are more suitable for low level feature representation. In our framework, the ROI pooling could be possibly applied on ``conv2'' to ``conv5'' and ``conv'' in Fig.~\ref{fig:framework}. In the experiment, a comparative study is made to show the distinctiveness of the feature extracted from these layers. In addition, we also test the possibility of concatenating features ROI-pooled from different stages. Features are \emph{l$_{2}$}-normalized before and after the concatenation. 

\subsection{Performance Enhancement}
\label{sec:boost}
In order to boost the performance of the proposed feature representation, the FCIS is modified in two aspects. Namely, the ResNet-101~\cite{He16}, upon which FCIS is built, is replaced by more powerful ResNeXt-101~\cite{Xie17}. In addition, to enable the network to be more robust to severe shape variations, deformable convolution~\cite{Dai17} is adopted in the last three bottle-neck blocks of ResNeXt-101.

\begin{figure}[htbp]
	\centering
	\includegraphics[width=0.86\linewidth]{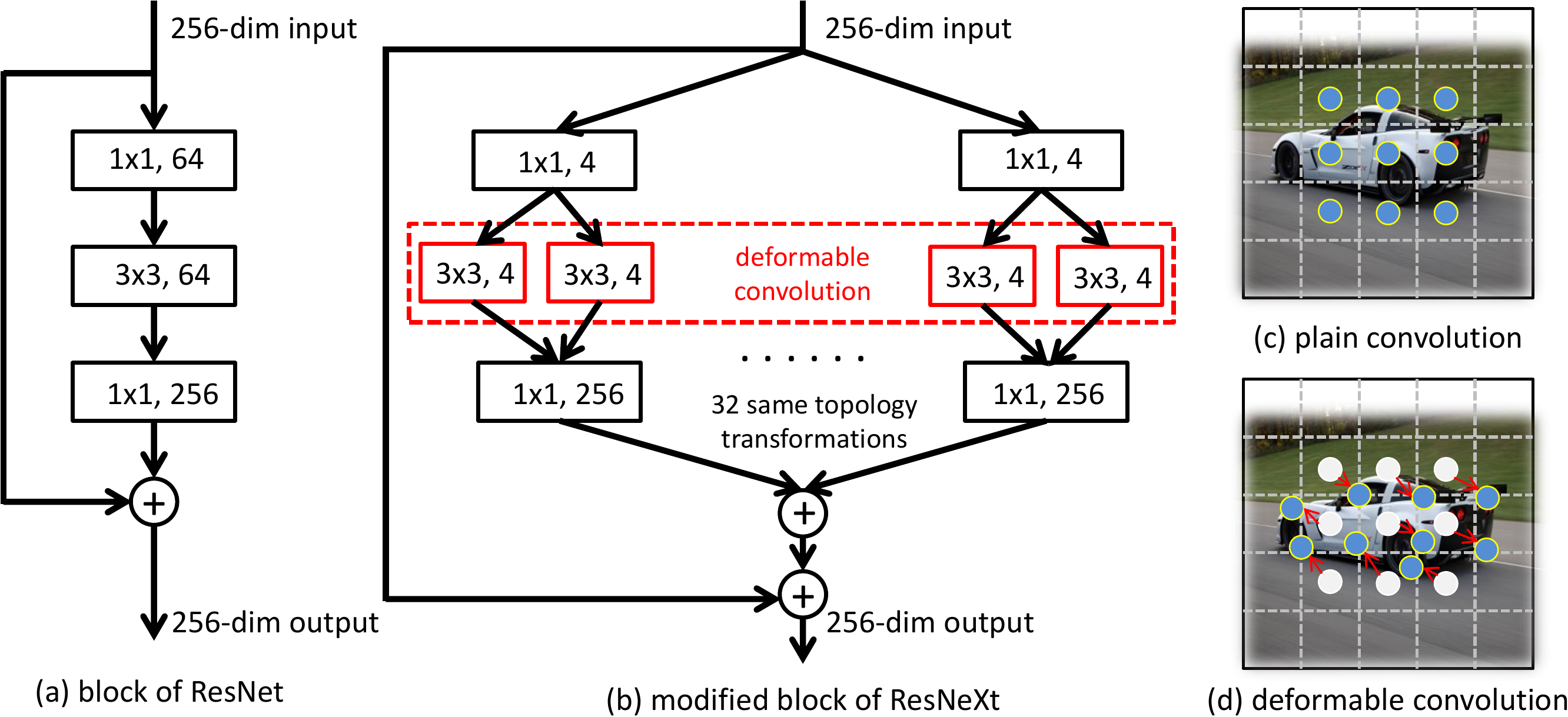}
	\caption{Comparison between ResNet and ResNeXt blocks. In figure (b), ResNeXt's block~\cite{Xie17} is embedded with deformable convolution~\cite{Dai17} with cardinality of \textit{32}. The size of filter and the number of filters are shown on each convolution layer. In the enhanced instance feature design, structure in (b) is adopted. The last \textit{3} bottle-neck blocks of ResNeXt-101 are replaced by deformable convolution given in figure (d).}
	\label{fig:blcok}
\end{figure}

As pointed out in~\cite{Li17}, the performance of ResNet~\cite{He16} gets saturated when its depth reaches to \textit{152}. To further improve the accuracy of this back-bone network, ResNet-101 is replaced by ResNeXt-101~\cite{Xie17} which corresponds to ``conv1-4'' and ``conv-5'' in Fig.~\ref{fig:framework}. Compared to ResNet, ResNeXt increases the \emph{cardinality} of the building blocks. Fig.~\ref{fig:blcok} show the difference between blocks of ResNet and ResNeXt. \emph{Cardinality} refers to the size of same-topology transformation aggregated in the building block. The cardinality of building blocks in our case is set to \textit{32}. This is to control the FLOPs complexity on the same level as ResNet. Similar as ResNet-101, the weights of the model are initialized from ImageNet~\cite{DengDSLL009} classification task. The layers (i.e., deformable convolution layer and RPN) absent from the pre-trained model are randomly initialized.

Visual instances usually undergo various irregular geometric transformations in real scenario, which causes heavy deformations in their appearances. Plain convolution modules in CNNs are inherently vulnerable to such kind of transformations. Inspired from~\cite{Dai17}, deformable convolutions are introduced to replace the plain convolution in the last three bottle-neck blocks of ResNeXt-101 to alleviate this problem~(illustrated in Fig. \ref{fig:blcok}). Fig.~\ref{fig:blcok}(d) shows the sampling structure of deformable convolution in contrast to plain one (Fig.~\ref{fig:blcok}(c)). The deformable convolution calculates a set of offsets for the ultimate sampling locations to better adapt to the deformations of the instance. The offsets are easily learned by applying a convolutional layer over the same input feature maps. As is revealed later in the experiments, both modifications proposed in this section boost the performance of instance segmentation and instance search.

\section{Evaluation Dataset Construction}
\label{sec:data}
Since the initiatives of instance search task in TRECVID~\cite{Awad17}, several instance search approaches have been proposed one after another over the past few years. However, the publicly available evaluation benchmark is slow to occur. Approaches~\cite{Tolias15,Kalantidis16, Salvador16} aiming for instance search are only evaluated on landmark datasets, typically Oxford5k~\cite{Philbin07}, Paris6k~\cite{Jegou08} and Holidays~\cite{Jegou08}. The evaluation does not reflect the real challenges, such as motion blur, partial occlusion, deformation and mutual object embedding, that instance search faces in the general cases. Dataset maintained by TRECVID~\cite{Awad17} avoids such kind of disadvantages, whereas it is only open to TRECVID participants. In this paper, a new dataset, namely \textit{Instance-160} is introduced. As visual object tracking and instance search are two similar tasks, \textit{Instance-160} is built based on the video sequences used for visual object tracking evaluation. On one hand, this avoids the painstaking efforts to annotate the instances from new video sequences. On the other hand, videos that are used for visual object tracking have been widely accepted benchmarks. The variety of variations and transformations that could happen on visual instances are incorporated.

\begin{figure}[htbp]
	\centering
	\subfigure[Eight sample queries in \textit{Instance-160}]
	{\includegraphics[width=0.72\linewidth]{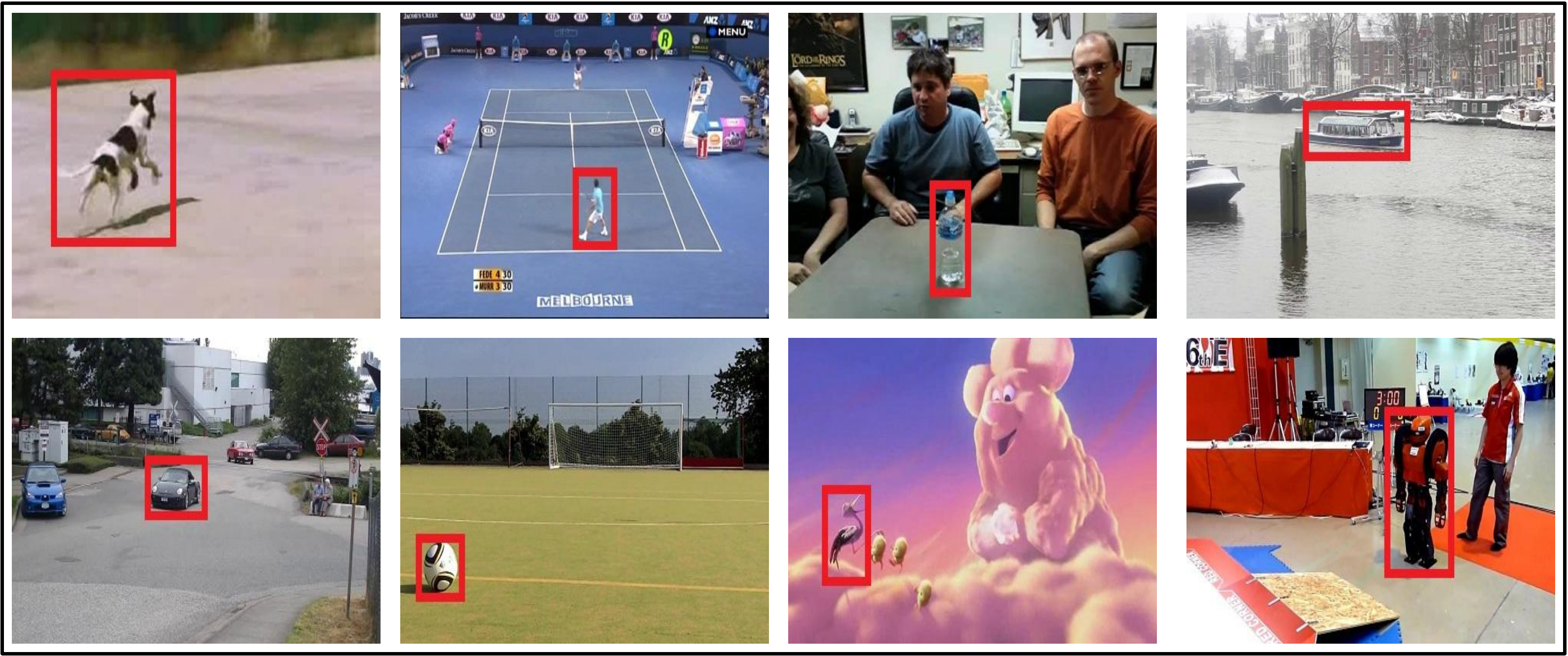}}
	\subfigure[Distribution of true-positive]
	{\includegraphics[width=0.72\linewidth]{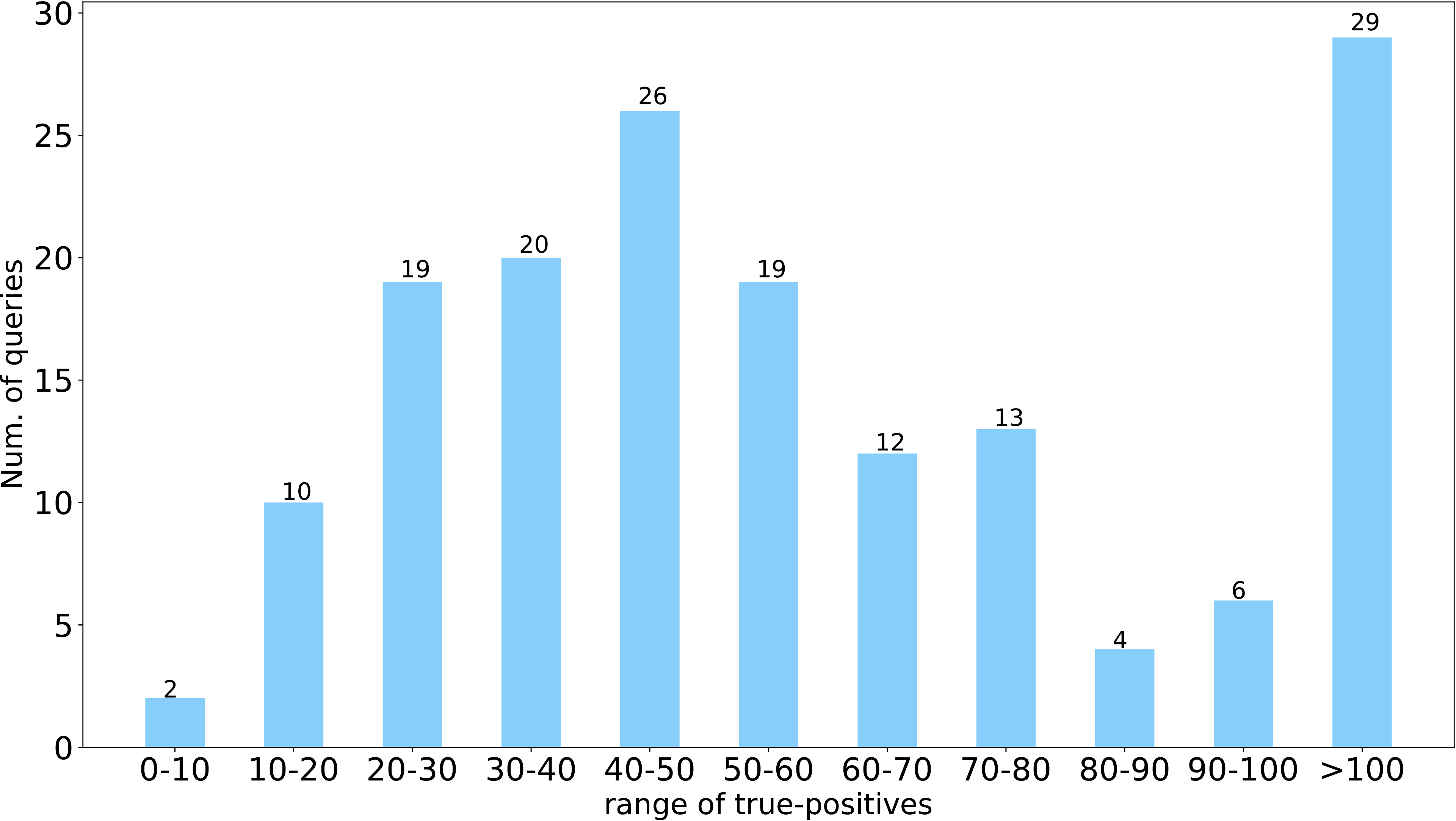}}
	\caption{Sample queries from \textit{Instance-160} and the number of true-positive statistics in \textit{Instance-160}.}
	\label{fig:query}
\end{figure}

In the object tracking, the tracking algorithm is required to track the target object (selected on the first frame) in the rest of video frames. In order to verify the robustness of the tracking algorithm, the test videos are collected from different scenarios and cover a wide range of objects. Most popular evaluation benchmarks are OTB2015~\cite{Wu15} and ALOV++~\cite{Smeulders14}. They are collected from diverse circumstance including illuminations, transparency, specularity, confusion with similar objects, clutter, occlusion, severe deformation, motion blur and low contrast. Since instance search arises from similar application scenarios as object tracking, the same challenges are seen in instance search. Nevertheless, it is worth noting that instance search is different from object tracking. The latter assumes the visual object varies following the temporal order. For this reason, the temporal information is more or less capitalized in various object tracking algorithms. While this is not the case for instance search. Moreover, the tracking algorithm is allowed to update the feature representation from time to time as the tracking continues. In contrast, feature representation, once has been designed, is fixed all the way in instance search.

When we construct \textit{Instance-160}, \textit{58} and \textit{102} sequences are selected from \textit{100} and \textit{300} video sequences from OTB2015 and ALOV++ respectively. The videos inside which the target instances are not covered by Microsoft COCO's \textit{80} categories are omitted. For each video, the first frame in which the query instance is given by a bounding box is extracted as the query side. For the rest, one frame is extracted for every other \textit{4} frames as the reference dataset. This results in \textit{11,885} reference images in total. Sample queries are seen in Fig. \ref{fig:query}(a). The distribution about the number of true-positives for all queries are shown in Fig. \ref{fig:query}(b). As shown in the figure, more than \textit{90\%} of the queries have more than \textit{20} true-positives for each.

\section{Experiments}
In this section, the proposed approach for instance search is evaluated on the dataset introduced in Section~\ref{sec:data}. Additionally, in order to verify the scalability of the presented approach, another \textit{1} million images randomly crawled from Flickr are incorporated as distractors. The performance evaluation is studied in comparison to several representative approaches. They are BoVW~\cite{Sivic03}, BoVW+HE~\cite{Jegou08}, R-MAC~\cite{Tolias15}, Deepvision~\cite{Salvador16} and CroW~\cite{Kalantidis16}. The last three are based on deep features. For each CNN-based method, the network is initialized with the default pre-trained model and configuration described in the corresponding paper. For BoVW and BoVW+HE, the same visual vocabulary sized of \textit{65,536} are used. The binary signature in HE is set to \textit{64} bits. The performance is measured by mAP at top-\textit{k}, where \textit{k} varies from \textit{10} to \textit{100}. This is due to the fact that more than \textit{95\%} the queries have more than \textit{10} corresponding true-positives as shown in Fig.~\ref{fig:query}(b).

Under the same training protocol introduced in~\cite{Li17}, the feasibility of the proposed enhancement strategies is validated on PASCAL VOC 2012~\cite{EveringhamGWWZ10}. Thereby, FCIS and FCIS in-planted with the proposed enhancement strategies are trained on Microsoft COCO 2014~\cite{LinMBHPRDZ14}. All the experiments are conducted on a workstation with four Nvidia Titan X GPUs and one \textit{3.20}GHz Intel CPU setup.

\subsection{Configuration Test on FCIS} 
Theoretically speaking, feature ROI-pooled from any layer could be used to represent the detected instance. The distinctiveness of these features varies from layer to layer. In the first experiment, the distinctiveness of instance-wise representation that are extracted from different layers is studied. The feature representation with the best distinctiveness (reflected by the highest mAP) is selected as the final feature representation. Additionally, we also investigate the possibility of concatenating features from different layers. 

\begin{figure}[htbp]
	\centering
	\includegraphics[width=0.84\linewidth]{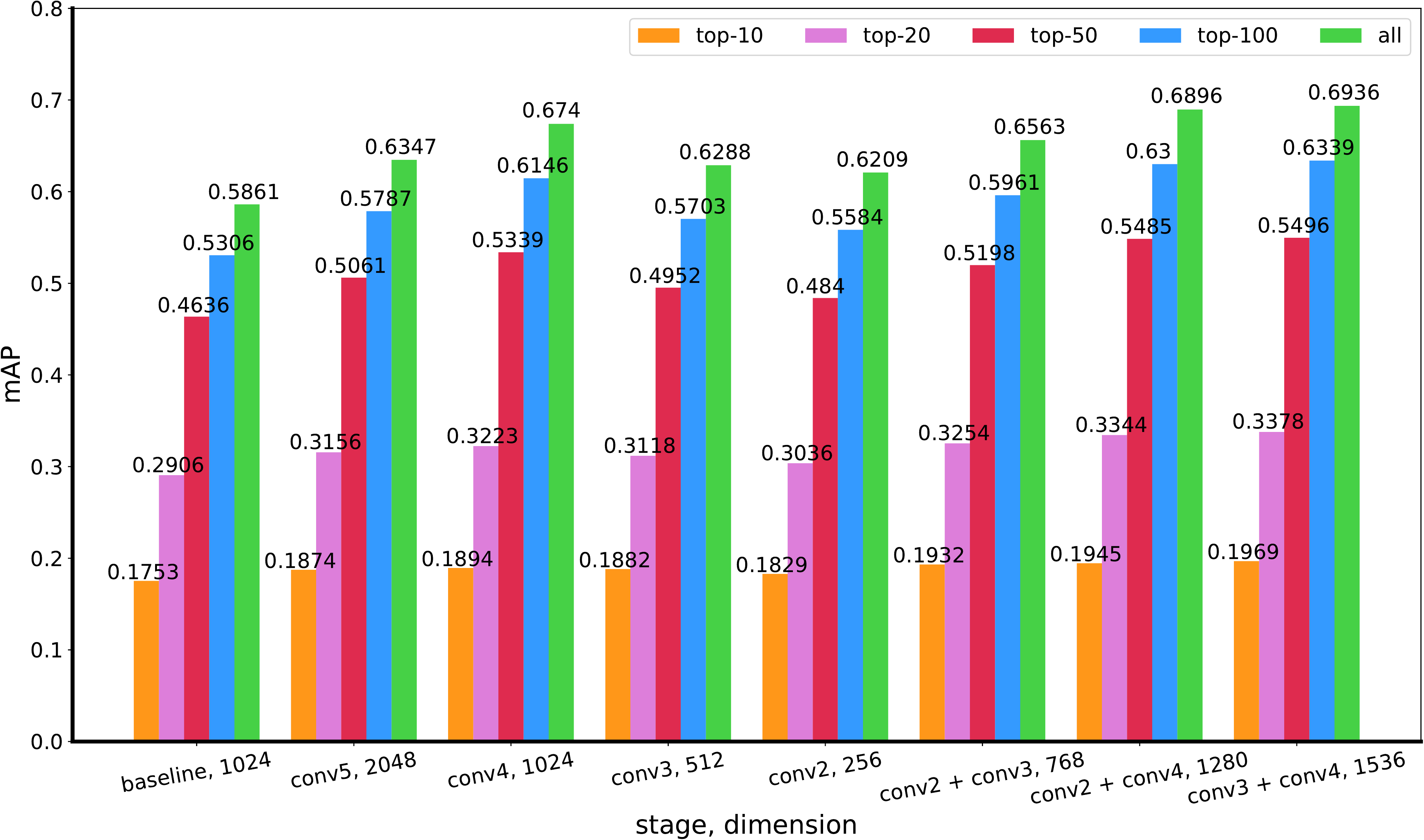}
	\caption{Performance of deep features extracted from different stages' convolution layer, including experiments with feature concatenation.}
	\label{fig:baseline}
\end{figure}

According to our observation, the category label for the segmented instance from FCIS is in high accuracy. It is therefore could be adopted for early pruning. Namely, the instance query only needs to compare with the candidate instances which share the same category label. Such pruning strategy speeds up the retrieval by two times without notable drop in mAP. In the following experiments, pruning scheme is adopted as default configuration for our approach.

In the first experiment, the distinctiveness of features from different layers of FCIS network is studied. We also investigate the performance of hybrid features that combining features from two layers. Feature derived from the ``conv'' (see Fig.~\ref{fig:framework}) layer is given as comparison baseline. 

Fig.~\ref{fig:baseline} summarizes the performance with features extracted from different stages. In the figure, mAP@\textit{10} and mAP@\textit{20} for all the configurations are low since not all potential true-positives are considered due to the fact that \textit{90\%} queries have more than \textit{20} true-positives (see Fig.~\ref{fig:query}(b)). As expected, features derived from intermediate layers perform better over feature from baseline (``conv''). Performance drops when features are derived from the shallow layers, such as ``conv2'' and ``conv3''. This basically indicates that it is sub-optimal to employ representations only kept with local visual patterns.  As seen in the figure, all three different combinations between features from different layers lead to better results. The combination of features from ``conv3'' and ``conv4'' achieves the best performance. As revealed in later experiment, this observation is consistent even we change the back-bone network from ResNet to ResNeXt.

\subsection{FCIS+XD versus FCIS}
In this section, we are going to investigate the performance achieved by two enhancement strategies proposed in Section~\ref{sec:boost}. Since FCIS is primarily designed for instance segmentation, the effectiveness of the enhanced FCIS network is studied first on instance segmentation task. In the experiment, the performance of FCIS with ResNeXt back-bone network and deformable convolution layer are studied both as separate runs and as a combination. FCIS supported with deformable convolution is denoted as FCIS+D. FCIS supported with ResNeXt-101 is denoted as FCIS+X. FCIS+XD denotes that FCIS powered by both enhancement strategies.

The performance evaluation is conducted on PASCAL VOC 2012~\cite{EveringhamGWWZ10} and Microsoft COCO 2014 test-dev~\cite{LinMBHPRDZ14}. $mAP^r@r$ is adopted for the evaluation. It basically calculates the mean of Average Precision (AP) measured for a method for which the corresponding recall exceeds \textit{r}. Notice that it is essentially different from mAP that we use to evaluate the instance search performance.

\begin{table}[htbp]
\centering
\scriptsize{
	\begin{tabular}{|l|p{1.5cm}<{\centering}|p{1.5cm}<{\centering}|}
		\hline
		Approach & mAP$^r$@0.5 & mAP$^r$@0.7\\
		\hline \hline
		FCIS     & 0.657  & 0.521\\	
		
		FCIS+D   & 0.667  & 0.528\\
		
		FCIS+X   & 0.658  & 0.526\\		
		
		FCIS+XD  & \textbf{0.675}  & \textbf{0.539}\\
		\hline
	\end{tabular}
	\caption{Performance comparison (measured by mAP$^r$) of FCIS with its variants on PASCAL VOC 2012~\cite{EveringhamGWWZ10}.}
\label{tab:segmentation_voc}
	}
\end{table}

\begin{table}[htbp]
\hspace{-0.15in}
\scriptsize{
	\begin{tabular}{|l|p{1.9cm}<{\centering}|p{1.2cm}<{\centering}|p{1.9cm}<{\centering}|p{1.9cm}<{\centering}|p{1.9cm}<{\centering}|p{1.9cm}<{\centering}|}
		\hline
		Approach & mAP$^r$@[0.5:0.95] & mAP$^r$@0.5  & mAP$^r$@[0.5:0.95] (small) & mAP$^r$@[0.5:0.95] (mid) & mAP$^r$@[0.5:0.95] (large)\\
		\hline \hline
		FCIS     & 0.292  & 0.495  & 0.071  & 0.313  & 0.500\\	
		
		FCIS+D   & 0.288  & 0.498  & 0.070  & 0.309  & 0.514\\
		
		FCIS+X   & 0.296  & 0.513  & 0.081  & 0.319  & 0.515\\		
		
		FCIS+XD  & \textbf{0.303} & \textbf{0.522}  & \textbf{0.082}  & \textbf{0.326}  & \textbf{0.528}\\
		\hline
	\end{tabular}
	\caption{Performance comparison (measured by mAP$^r$) of FCIS with its variants on Microsoft COCO 2014 test-dev~\cite{LinMBHPRDZ14}.}\label{tab:segmentation_coco}
	}
\end{table}

The performance of instance segmentation using FCIS and its variants is summarized in Table~\ref{tab:segmentation_voc} and Table~\ref{tab:segmentation_coco}. On the two datasets PASCAL VOC 2012 and Microsoft COCO 2014, both networks individually supported by deformable convolution layers and ResNeXt bottle-neck blocks (denoted as FCIS+D and FCIS+X respectively) are able to achieve better results in comparison to original FCIS architecture. When both of these enhancement strategies are adopted (given as FCIS+XD), the best segmentation accuracy is attained. As we verified on PASCAL VOC 2012 and Microsoft COCO 2014, the segmentation accuracy of FCIS is relatively improved by \textit{2.7\%} and \textit{5.5\%} measured with mAP$^r$@0.5 respectively by FCIS+XD. Such results indicate that the enhancement strategies proposed in Section~\ref{sec:boost} are all effective in boosting the performance of instance segmentation task. 

\begin{figure}[htbp]
	\centering
	\includegraphics[width=0.78\linewidth]{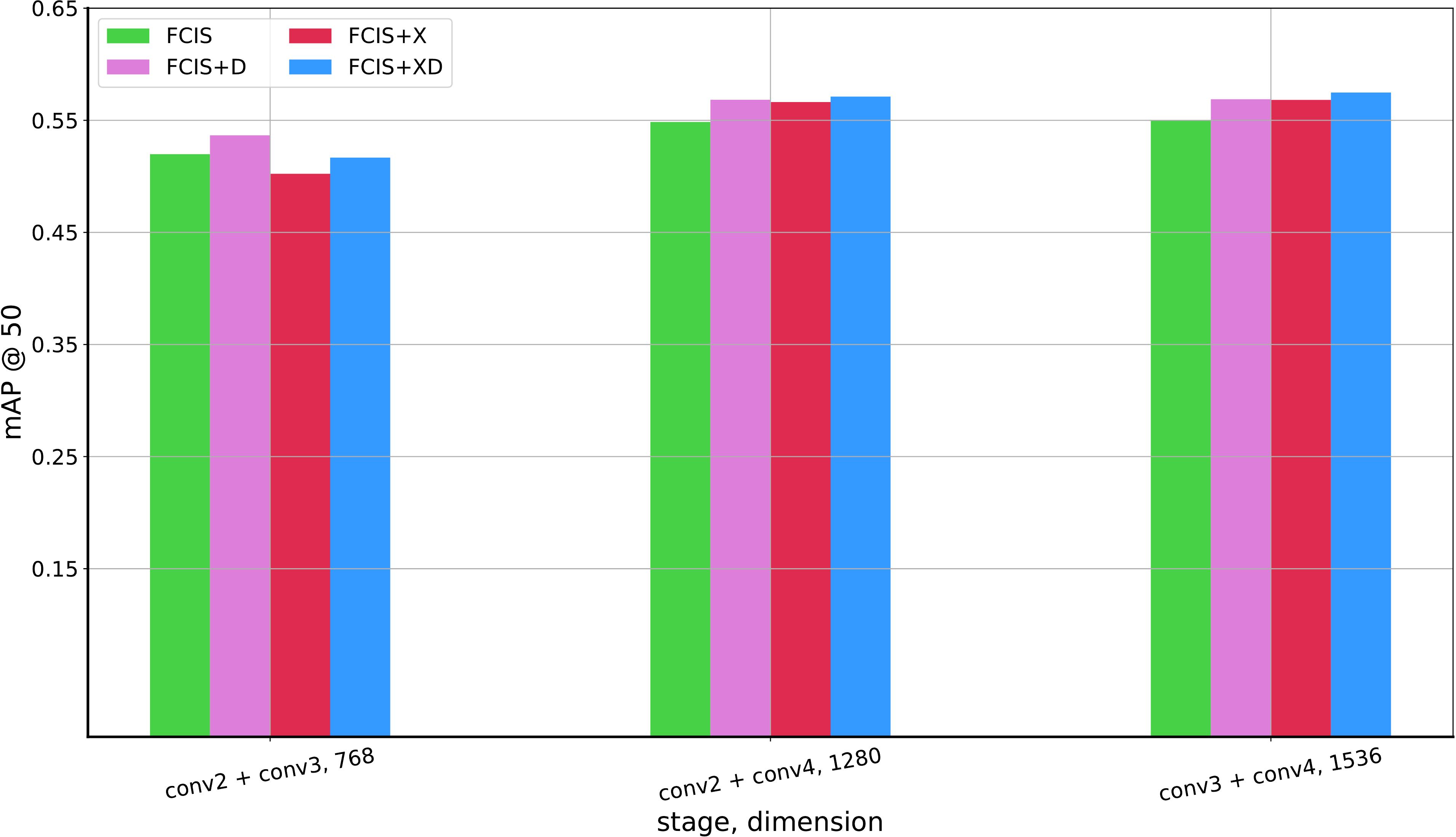}
	\caption{Performance of FCIS, FCIS+D, FCIS+D and FCIS+XD with the hybrid features from different layers. The performance is  measured by mAP at top-\textit{50} on \textit{Instance-160}.}
	\label{fig:modification}
\end{figure}

In addition, we further study the performance of the features derived from FCIS+D, FCIS+X and FCIS+XD when they are adopted for instance search task. Similar as FCIS, the other three networks are trained on Microsoft COCO 2014~\cite{LinMBHPRDZ14}. Fig.~\ref{fig:modification} presents the performance of FCIS and its variants on \textit{Instance-160}. mAPs at top-\textit{50} are presented. Since hybrid features from different layers are always better than the ones from single layer, the results of features derived from single layer are omitted. As seen from the figure, hybrid features from ``conv3 + conv4'' achieve the best result. This is consistent with the observation on the results shown in Fig.~\ref{fig:baseline}. In the following experiments, hybrid feature from ``conv3'' and ``conv4'' is selected as the feature representation for each detected instance. 

Table~\ref{tab:self-cmpr} further shows the performance of FCIS and its enhancements on \textit{Instance-160}. For all different networks, the features are extracted from ``conv3'' and ``conv4''. Due to the high accuracy on instance level segmentation, superior performance is observed with FCIS+XD across all the rankings. It outperforms FCIS by a constant \textit{2-3\%} margin. In the rest of our experiments, FCIS with ResNeXt back-bone network and deformable convolution, namely FCIS+XD is selected as the standard configuration for our approach.



\begin{table}[htbp]
\centering
\scriptsize{
	\begin{tabular}{|l|c|p{0.8cm}<{\centering}|p{1.0cm}<{\centering}|p{1.0cm}<{\centering}|p{1.0cm}<{\centering}|p{1.0cm}<{\centering}|}
		\hline
		Approach & Dist. measure   & top-10   & top-20  & top-50  & top-100 & all\\
		\hline\hline
		FCIS     & \textit{Cosine} & 0.1969   & 0.3378  & 0.5496  & 0.6339  & 0.6936\\
		\hline
		FCIS+D   & \textit{Cosine} & 0.2089   & 0.3517  & 0.5688  & 0.6535  & 0.7127\\
		
		FCIS+X   & \textit{Cosine} & 0.2078   & 0.3522  & 0.5682  & 0.6537  & 0.7125\\
		
		FCIS+XD  &\textit{Cosine}  & \textbf{0.2109} & \textbf{0.3558} & \textbf{0.5747} & \textbf{0.6585} & \textbf{0.7237}\\
		\hline
	\end{tabular}
	\caption{Performance (mAP) of FCIS, FCIS+D, FCIS+X and FCIS+XD with the hybrid features of ``conv3+conv4'' on \textit{Instance-160}.}\label{tab:self-cmpr}
	}
\end{table}

\subsection{Comparison to State-of-the-art Approaches}
In this section, the performance of proposed FCIS+XD is studied in comparison to five representative approaches in the literature. They are two local feature based approaches BoVW~\cite{Sivic03} and BoVW+HE~\cite{Jegou08} and three deep feature based approaches R-MAC~\cite{Tolias15}, Deepvision~\cite{Salvador16} (denoted as DV-Vgg) and CroW~\cite{Kalantidis16}. For Deepvision, the search is carried out in two steps. In the first step, the top-ranked candidates are produced by image level comparison. In the second step, instance level search is carried out on the top-\textit{100} candidates. In order to make a more fair comparison between Deepvision and our approach, another run is also conducted for Deepvision. In this new run, back-bone network of Deepvision is replaced by ResNet-101, which becomes the same as FCIS. The filtering scheme in the first step is disabled. This run is denoted as DV-Res.

\begin{table}[htbp]
\centering
\scriptsize{
	\begin{tabular}{|l|c|p{0.8cm}<{\centering}|p{1.0cm}<{\centering}|p{1.0cm}<{\centering}|p{1.0cm}<{\centering}|p{1.0cm}<{\centering}|}
		\hline
		Approach                     &Dist. measure     & top-10 & top-20 & top-50 & top-100 & all\\
		\hline\hline
		BoVW~\cite{Sivic03}          &\textit{Cosine}   &0.1061  &0.1651  &0.2483  &0.2806  &0.3141\\
		
		BoVW+HE~\cite{Jegou08}       &\textit{Cosine}   &0.1483  &0.2359  &0.3553  &0.4033  &0.4375\\
		
		R-MAC~\cite{Tolias15}        &\textit{Cosine}   &0.1014  &0.1685  &0.2680  &0.3071  &0.3577\\
		
		CroW~\cite{Kalantidis16}     & $\textit{l}_2$   &0.0733  &0.1296  &0.2391  &0.2840  &0.3375\\		
		\hline		
		
		DV-Res~\cite{Salvador16} &\textit{Cosine}   &0.1763  &0.2908  &0.4609  &0.5239  &0.5790\\		
		
		DV-Vgg~\cite{Salvador16}     &\textit{Cosine}   &0.1939  &0.3282  &0.5413  &0.6660  &0.7306\\		
		
		\hline
		FCIS+XD                      &\textit{Cosine}	&\textbf{0.2109}  &\textbf{0.3558}  &\textbf{0.5747}  &0.6585  &0.7237\\
		\hline
	\end{tabular}
	\caption{Performance (mAP) of FCIS+XD compared to five representative approaches in the literature.}\label{tab:comparison}
	}
\end{table}

Table~\ref{tab:comparison} shows the performance from all approaches. As seen from the table, DV-Vgg and FCIS+XD show considerably better performance than the rest. BoVW+HE still shows competitive performance in comparison to deep feature approaches such as R-MAC and CroW. Although the results from Deepvision are very close to FCIS+XD, they do not reflect real behavior of Deepvision. In \textit{Instance-160}, the videos are primarily collected from visual tracking evaluation. In many cases, the query instance shares similar background scene as the reference images. So that true instances are retrieved by Deepvision due to their similar background. For this reason, the image-wise feature representation in Deepvision still works seemingly well. However, the performance of Deepvision drops considerably when the target instances are cluttered in different backgrounds. This will be confirmed by another experiment afterwards. Another disadvantage for Deepvision lies in its low accuracy of generated instance bounding box. As shown in the table, the mAP of DV-Res is even lower than original FCIS (see Table~\ref{tab:self-cmpr}) although it is already powered by ResNet. This is mainly caused by its imprecise feature representation of each instance. In contrast, FCIS+XD is able to generate precise instance-level bounding-boxes owing to its precise object category-level classification and pixel-level mask prediction.


\begin{table}[htbp]
\centering
\scriptsize{
	\begin{tabular}{|l|c|p{1.2cm}<{\centering}|p{1.08cm}<{\centering}|p{1.08cm}<{\centering}|p{1.08cm}<{\centering}|p{1.08cm}<{\centering}|}
		\hline
		Approach & Dist. measure & top-10    & top-20 & top-50 & top-100 & all\\
		\hline \hline
		DV-Vgg~\cite{Salvador16} & \textit{Cosine}  & 0.1925 & 0.2979  & 0.4642 & 0.5585 & 0.5894\\
		\hline
		FCIS       & \textit{Cosine}  & 0.2521 & 0.4075  & 0.6094 & 0.6582 & 0.6975\\
		FCIS+XD  	& \textit{Cosine} & \textbf{0.2624} & \textbf{0.4301} & \textbf{0.6467} & \textbf{0.6975} & \textbf{0.7366}\\
		\hline
	\end{tabular}
	\caption{Performance comparison (measured by mAP) of our method to Deepvision on \textit{40} queries in which heavy background variations are observed.}\label{tab:subset}
	}
\end{table}

In order to further confirm our observation about Deepvision, \textit{40} queries from \textit{Instance-160}, in which severe background variations are observed, are selected to verify its real behavior. Table \ref{tab:subset} shows the performance of FCIS, FCIS+XD and Deepvision on \textit{40} queries. As observed from the table, the performance of Deepvision drops considerably compared to that of Table~\ref{tab:comparison}. As the background scenes from the instance query and the reference images are dissimilar, the first round search in Deepvision becomes ineffective since it is based on image-wise feature. As the consequence, decent results are not expected from the re-ranking stage since many true-positives are already missed in the first stage. Another disadvantage of this approach is that one has to keep two types of features. One is on image level, another is on region level, which induce heavy computational overhead.

\subsection{Scalability Test}

\begin{figure}[htbp]
	\centering
	\includegraphics[width=0.8\linewidth]{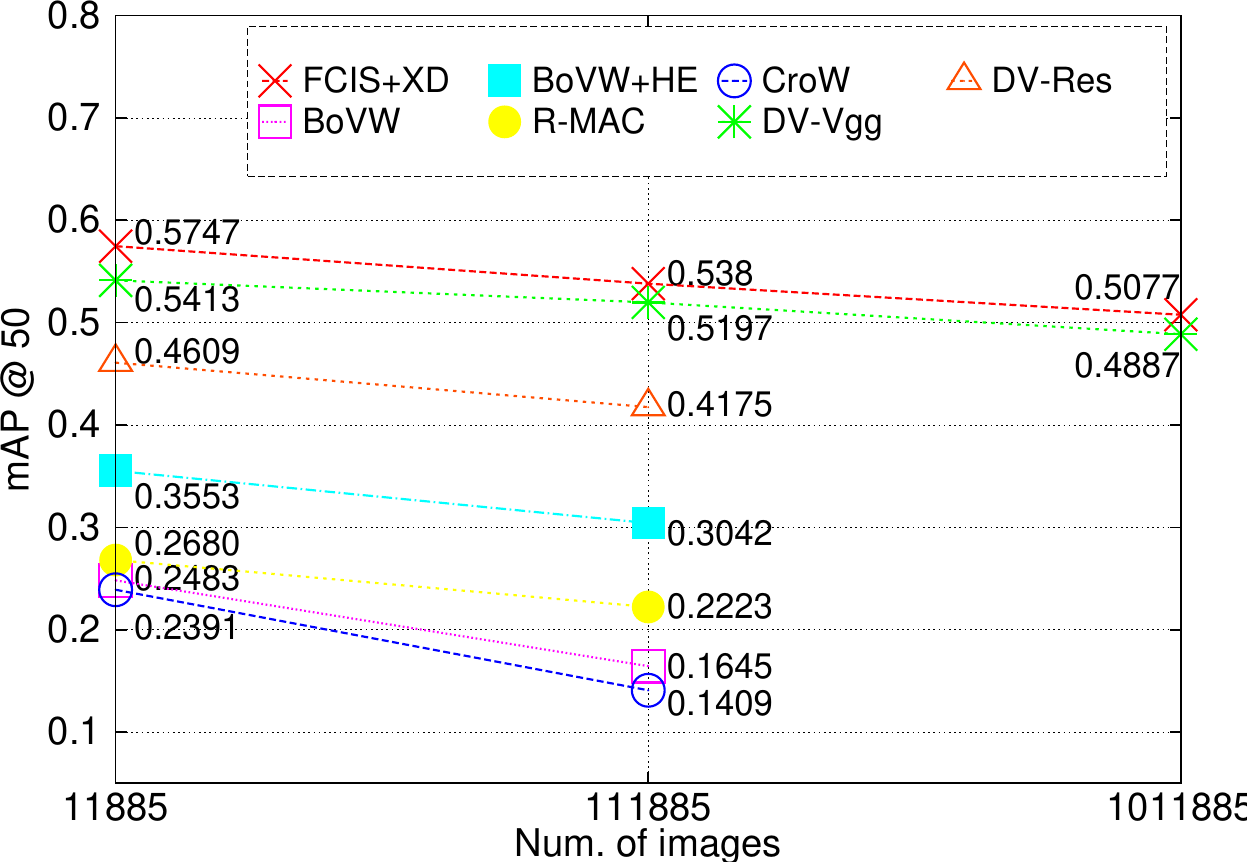}
	\caption{Scalability test in comparison to five state-of-the-art approaches. The performance is measured by mAP at top-\textit{50}.}
	\label{fig:distract}
\end{figure}

In this section, the scalability of the proposed feature representation is studied. In the experiment, \textit{1} million distractor images are added in the reference set. The same processing pipeline is undertaken on this \textit{1} million images. In the experiment, five representative approaches are considered. For FCIS+XD, \textit{1,648,654} instances are extracted from the distractor images, each of which is represented as an \textit{1,536}-dimensional feature vector.

\begin{figure}[htbp]
	\centering
	\includegraphics[width=0.98\linewidth]{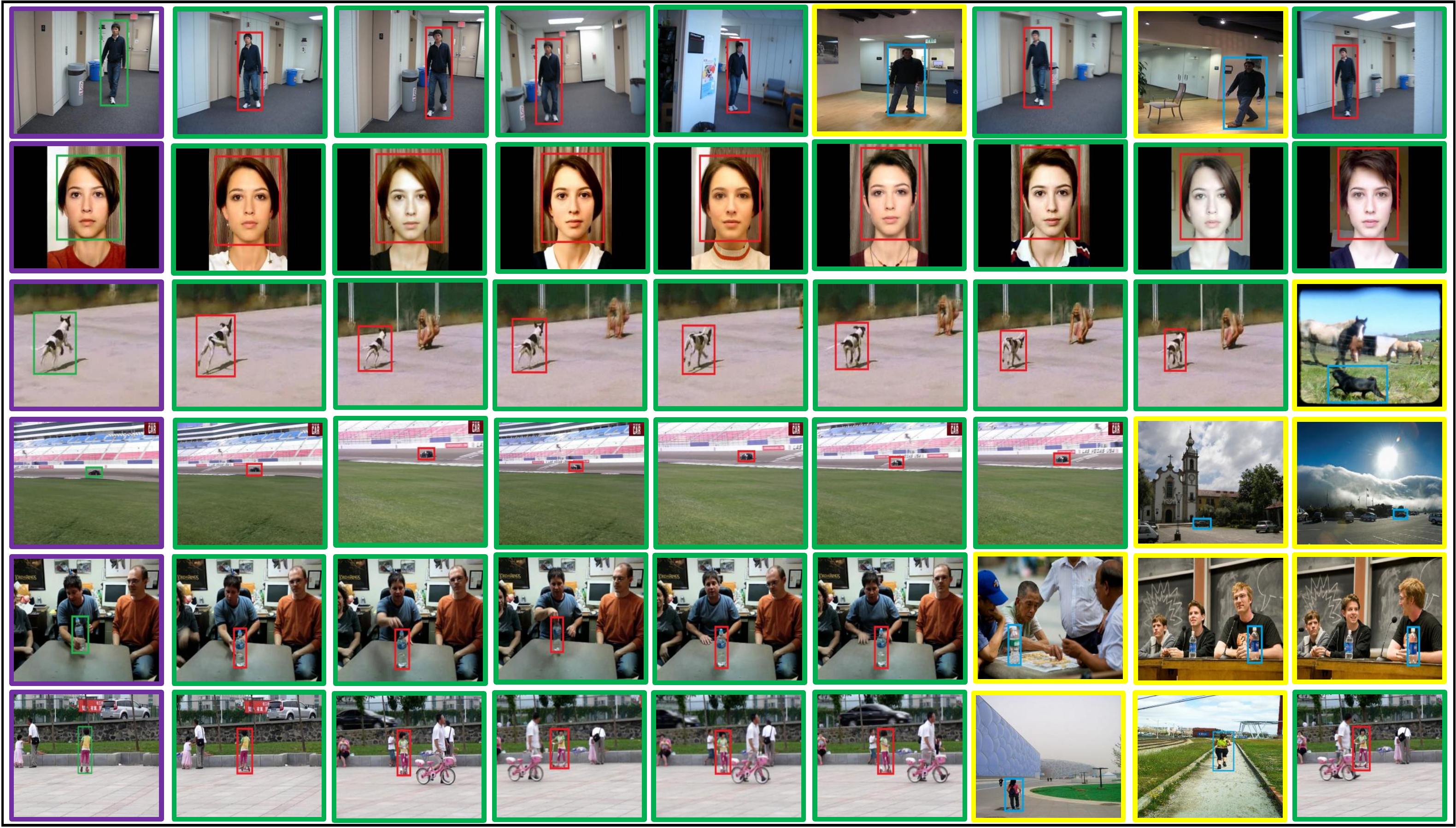}
	\caption{Top-\textit{8} search results of six sample queries produced by FCIS+XD with \textit{1} million distractor images.}
	\label{fig:result}
\end{figure}

As seen from Fig.~\ref{fig:distract}, FCIS+XD shows the best scalability. It outperforms Deepvision by a constant margin. As the computation cost is high and the results from BoVW, BoVW+HE, R-MAC, CroW and DV-Res are already much poorer than FCIS+XD and Deepvision (DV-Vgg) with 100K distractors, further verification on the whole \textit{1} million distractors is not carried out for these approaches. Fig.~\ref{fig:result} shows six instance search results produced by FCIS+XD. As shown in the figure, all the top-\textit{8} results for each individual query are meaningful. Although a few false-positive instances are returned, they indeed exhibit very close appearance as the query.


\section{Conclusion}
We have presented a promising way of instance level feature representation for instance search. This representation is built upon a fully convolutional network that is originally used for instance segmentation. With the precise instance segmentation, the feature is derived by ROI pooling on the feature maps. To further boost its performance, two enhancement strategies are proposed. The distinctiveness and scalability of this feature have been comprehensively studied. As shown in the experiment, it outperforms most of the representative approaches in the literature. Considering the lack of publicly available evaluation benchmark, a medium-scale dataset for instance search is introduced by harvesting videos from object tracking benchmarks. Currently, the types of instances that our approach could handle with are restricted to Microsoft COCO-\textit{80} categories. Although it already covers variety of instances that we encounter in the daily life, exploring more generic instance segmentation model that works beyond \textit{80} categories will be our future research focus.

\section{Acknowledgments}
This work is supported by National Natural Science Foundation of China under grants 61572408 and grants of Xiamen University 0630-ZK1083.

\bibliographystyle{elsarticle-num}
\bibliography{yzhan}



\end{document}